# Vagueness of Linguistic variable

Supriya Raheja, Smita Rajpal

**Abstract**— In the area of computer science focusing on creating machines that can engage on behaviors that humans consider intelligent. The ability to create intelligent machines has intrigued humans since ancient times and today with the advent of the computer and 50 years of research into various programming techniques, the dream of smart machines is becoming a reality. Researchers are creating systems which can mimic human thought, understand speech, beat the best human chessplayer, and countless other feats never before possible. Ability of the human to estimate the information is most brightly shown in using of natural languages. Using words of a natural language for valuation qualitative attributes, for example, the person pawns uncertainty in form of vagueness in itself estimations. Vague sets, vague judgments, vague conclusions takes place there and then, where and when the reasonable subject exists and also is interested in something. The vague sets theory has arisen as the answer to an illegibility of language the reasonable subject speaks. Language of a reasonable subject is generated by vague events which are created by the reason and which are operated by the mind. The theory of vague sets represents an attempt to find such approximation of vague grouping which would be more convenient, than the classical theory of sets in situations where the natural language plays a significant role. Such theory has been offered by known American mathematician Gau and Buehrer .In our paper we are describing how vagueness of linguistic variables can be solved by using the vague set theory.This paper is mainly designed for one of directions of the eventology (the theory of the random vague events), which has arisen within the limits of the probability theory and which pursue the unique purpose to describe eventologically a movement of reason.

———————— ◆ ————————

## 1 INTRODUCTION

Theory of Vague Sets  [1]
First of all we recollect basic preliminaries of vague theory. There are a number of generalizations [1, 2, 3, 9, 10]   of Zadeh's fuzzy set theory [2]  so far reported in the literature viz., i-v fuzzy theory, two-fold fuzzy theory, vague theory, intuitionistic fuzzy theory, probabilistic fuzzy theory, L-fuzzy theory, etc.  The notion of vague theory recently introduced in IEEE by Gau and Buehrer [1] is of interest to us for this present work.   For each such generalization, one (or more) extra edge is added with the fuzzy theory with specialized type of aim and objective.  Thus, a number of higher order fuzzy sets are now in literatures and are being applied into the corresponding more specialized application domains.  While fuzzy sets are applicable to each of such application domains, higher order fuzzy sets can not, because of their specialization in character by birth.  Application of higher order fuzzy sets makes the solution-procedure more complex, but if the complexity on computation-time, computation-volume or memory-space are not the matter of concern then a better results could be achieved.  Vague sets defined recently by Gau and Buehrer [1]  have also an extra edge over fuzzy sets.  Let U be a universe, say the collection of all students of Calcutta High School.  Let A be a vague set of all "good-in-maths students" of the universe U,   and B be a fuzzy set of all "good-in-maths students" of U.  Suppose that an intellectual Manager $M_1$ proposes the membership value $\mu_B(x)$ for the element x in

• *Supriya Raheja,  ITM University, Gurgaon, Haryana*

• *Smita Rajpal, ITM University, Gurgaon,Haryana*

the fuzzy set B by his best intellectual capability.  On the contrary, another intellectual Manager $M_2$ proposes independently two membership values $t_A(x)$ and $f_A(x)$ for the same element in the vague set A by his best intellectual capability. The amount $t_A(x)$ is the true-membership value of x and $f_A(x)$ is the false-membership value of x in the vague set A.  Both $M_1$ and $M_2$ being human agents have their limitation of perception, judgment, processing-capability with real life complex situations. In the case of fuzzy set B, the manager $M_1$ proposes the membership value $\mu_B(x)$  and proceeds to his next computation. There is no higher order check for this membership value in general.  In the later case, the manager $M_2$ proposes independently the membership values $t_A(x)$    and $f_A(x)$, and makes a check at this base-point itself by exploiting the constraint   $t_A(x) + f_A(x) \le 1$.  If it is not honored, the manager has a scope to rethink, to reshuffle his judgment procedure either on 'evidence against' or on 'evidence for' or on both.  The two membership values are proposed independently, but they are mathematically not independent. They are mathematically constrained. This is the breaking philosophy in Gau and Buehrer's theory vague sets [9].

In his classical work [2], Zadeh proposed the theory of fuzzy sets.  Since then it has been applied in wide varieties of fields like Computer Science, Management Science, Medical Sciences, Engineering problems etc. to list a few only.

 Let   U  =  {$u_1$, $u_2$…, $u_n$}    be the universe of discourse. The membership function for fuzzy sets can have functional values from the closed interval [0, 1].  A Fuzzy set **A** in U is defined as the set of ordered pairs  **A**  =  { ( u, $\mu_A(u)$ ) :  u $\in$ U  }, where $\mu_A(u)$ is the grade of membership of the element u in the set **A**.  The greater $\mu_A(u)$, the greater is the truth of the statement that 'the element u belongs to the set **A'**. With this philosophy, Prof. Zadeh generalized the notion of crisp subset of classical set theory.

But Gau and Buehrer [1] pointed out that this single value combines the 'evidence for u' and the 'evidence against u'. It does not indicate the 'evidence for u' and the 'evidence against u', and it does not also indicate how much there is of each. Consequently, there is a genuine necessity of a model like vague sets, kind of higher order fuzzy sets, which could be treated as a generalization of Zadeh's fuzzy sets [2].

## 2 DEFINITION

A vague set (or in short VS) A in the universe of discourse U is characterized by two membership functions given by:-

(i) A truth membership function

$$t_A : U \rightarrow [0, 1],\quad \text{and}$$

(ii) A false membership function

$$f_A : U \rightarrow [0, 1],$$

Where $t_A(u)$ is a lower bound of the grade of membership of u derived from the 'evidence for u', and $f_A(u)$ is a lower bound on the negation of u derived from the 'evidence against u', and their total amount can not exceed 1 i.e. $t_A(u) + f_A(u) \leq 1$.

Thus the grade of membership of u in the vague set A is bounded by a subinterval $[t_A(u), 1- f_A(u)]$ of $[0,1]$. This indicates that if the actual grade of membership is $\mu(u)$, then $t_A(u) \leq \mu(u) \leq 1- f_A(u)$.

The vague set A is written as A = { < u, [$t_A(u)$, $f_A(u)$] > : u ∈ U }, where the interval $[t_A(u), 1- f_A(u)]$ is called the 'vague value' of u in A and is denoted by $V_A(u)$.

For example, consider a universe U = {DOG, CAT, and RAT}. A vague set A of U could be A = {<DOG, [.7,.2]>, <CAT, [.3,.5]., <RAT, [.4,.6]> }.

*It is worth to mention here that interval-valued fuzzy sets (i-v fuzzy sets)* [2] *are not vague sets.* In i-v fuzzy sets, an interval valued membership value is assigned to each element of the universe considering the 'evidence for u' only, without considering 'evidence against u'. In vague sets both are independently proposed by the decision maker. This makes a major difference in the judgment about the grade of membership.

**Linguistic Variable**

One of the fundamental concepts entered by Zadeh is the concept of a linguistic variable [2, 3]. It is the five objects,

$$\langle L, T(L), U, G, S \rangle, \tag{1}$$

where *L* is a name of a variable; *T(L)* is a set of its values (a term-set), being syntagmas[1]; *U* is an universal set; *G* is a syntactic rule (it is a context-free grammar more often), using which we can form syntagmas $A, B,… \in T(L)$; *M* is a semantic rule, using which to each syntagma $A \in T(L)$ is attributed its value, being by vague set in universal set *U*. Syntactic procedure *G* allows not only to operate with elements of the term-set *T(L)*, but also to generate new syntagmas (terms) by means of links "and", "or", negation "not", linguistic gradation, such as "very much", " more or less ", "essentially", etc. A semantic rule *S* defines a way of calculation of sense of any term from the set *T(L)*. Gau and Buehrer has suggested to formalize atomic terms by vague sets Ã in *U*. Links, uncertainty and negation are treated as operators which alter sense of primary terms in the special way independent of a context [2]. Now in the theory of vague sets for formalization of links "and", "or" t-norms and t-conorm are used widely. These operations are well enough studied and underlie of many formal constructions of vague logic.

---

[1] Any part of the offer making sense and formed according to grammatical rules is a syntagma.

Gau and Buehrer's theory is the most successfully applied there and then, where and when the vagueness is generated by presence of the person and of its reason. This problem is the main for one of directions of the eventology (the theory of the random vague events), which has arisen within the limits of the probability theory and which pursues the unique purpose to describe eventologically a movement of the reason. Enentology is a theory which unexpected for a foreign sight applies for creation of original and rather natural mathematical language for discussion of the general theoretical bases of uncertainty. The eventological substantiation of the theory of vague sets of Gau and Buehrer was offered by O.J.Vorob'ov in 2004. General principles of the theory of fuzzy events using the eventological language are stated in [4].

Here the eventological formalization of a linguistic variable is offered.

The five objects,

$$\langle L, X, \Omega, G_E, S_E \rangle, \tag{2}$$

is called an eventological linguistic variable (E-linguistic variable). *L* is a name of variable; *X* is a set of names of events; $\Omega$ is a set of elementary events; $G_E$ is a syntactic rule (context-free grammar); $S_E$ is an eventological semantic rule.

It is necessary to notice, that the set of names of events *X* is equivalent to the set of atomic syntagmas *T(L)* from (1), the set of elementary events and $\Omega$ coincides with the universal set *U*, the context-free grammar *G* and the $G_E$ are equivalent. The essential distinction between (1) and (2) consists in the definition of the semantic rule $S_E$ based on concept of vague experiment.

Two phenomena, namely uncertainty and a vagueness which scientific importance has increased mainly in the last century are discussed in [3] in detail enough. Both these concepts characterize situations in which we consider the phenomena surrounding us. They concern to volume of the knowledge having (or able to be) in our disposal which, however are limited. The phenomenon of uncertainty arises because of lack of the knowledge concerning occurrence of some event. It meets till the moment of carrying out of some experiment which result is unknown for us. The mathematical model of the pheno-menon of uncertainty is based on the device of probability theory. The vagueness concerns to a way of the description of the event and does not consider a question about it appearance. The mathematical model of the phenomenon of the vagueness is based on the device of the theory of vague sets. In [3] it is noted, that these phenomena are represented like two sides supplementing each other of the most general phenomenon which authors name *indeterminacy*. From our point of view indeterminacy, considered in [3] is very well described by the fuzzy experiment offered in work by O.Yu.Vorob'ov [4].

A vague experiment is inconceivable without participation of set of the individual reasonable subjects making an integral part of vague experiment. Each individual reasonable subject is the participant of vague experiment and has his own opinion on, whether it is possible to characterize event in vague experiment by the given syntagma or not. All of their opinions form *vague event* as an outcome of vague experiment.

Let's illustrate formalization of a E-linguistic variable "*Age*". Let's consider vague experiment in which participate $M$ reasonable subjects. Each reasonable subject $\mu \in M$ is characterized by syntagmas (names) from set $X$. $X=\{x, y\}$, where $x$ is "*the young man*", $y$ is "*the young woman*" for example. Then the intercept $[0, 80]$ of the real axis acts as a set of elementary events $\Omega$. The age of the concrete person corresponds to an elementary event $\omega \in \Omega$. Let $(\Omega, F, P)$ is probabilistic space. Let's define a *matrix of the selected random events* [4] as a set of events $X_M = \{x_\mu, x \in X, \mu \in M\}$, where $x_\mu \in F$ are measurable random events concerning algebra. Thus, each pair $(x,\mu) \in X \times M$ defines one random event $x_\mu \subseteq \Omega$. In our example random event $x_\mu = [a_\mu, b_\mu]$ is a judgment of reason $\mu$, in which it correlates an age intercept $[a_\mu, b_\mu] \subseteq \Omega$ to a syntagma (name) (tab. 1).

Tab.1. Matrix of the selected events

| X | ... | $[a_\lambda, b_\lambda]$ | ... | $[a_\mu, b_\mu]$ | ... |
|---|---|---|---|---|---|
| Y | ... | $[c_\lambda, d_\lambda]$ | ... | $[c_\mu, d_\mu]$ | ... |
| | ... | $\lambda$ | ... | $\mu$ | ... |

The eventology [4] defines *M*-vague event as set of usual Kolmogorov's events, when $\mu \in M$. Thus, sets of elements of rows of a matrix of selected events $x_M = \{x_\mu, \mu \in M\}$ define vague events $\widetilde{x} = x_M$ and form the set of *M*-vague events $\widetilde{X} = \{\widetilde{x}, x \in X\}$. In our example we consider a set of *M*-vague events $\widetilde{X} = \{\widetilde{x}, \widetilde{y}\}$, generated by the set $X$, where $\widetilde{x} = \{[a_\mu, b_\mu], \mu \in M\}$ is "*the young man*" and $\widetilde{y} = \{[c_\mu, d_\mu], \mu \in M\}$ $\mu \in M$ "*the young woman*".

In the Gau and Buehrer's theory the vague set is defined by membership function on $U$, which kind cannot be deduced theoretically from more simple concepts, and it is established in each problem, proceeding from external in relation to the theory of consideration. Eventology unequivocally defines *eventological membership function* of vague event as the indicator of vague event

$$\mathbf{1}_{\widetilde{x}}(\omega) = \frac{1}{|M|} \sum_{\mu \in M} \mathbf{1}_{x_\mu}(\omega), \ x \in X,$$

(3)

Where $\mathbf{1}_{x_\mu}(\omega)$ - indicators of "usual" events $x_\mu$ which take sets of events $\widetilde{x}$. In our example (3) for corresponding vague events from the $X$ it will be transformed to a kind (fig. 1):

$$\mathbf{1}_{\widetilde{x}}(\omega) = \frac{1}{|M|} \sum_{\mu \in M} \mathbf{1}_{[a_\mu, b_\mu]}(\omega),$$

$$\mathbf{1}_{\widetilde{y}}(\omega) = \frac{1}{|M|} \sum_{\mu \in M} \mathbf{1}_{[c_\mu, d_\mu]}(\omega).$$

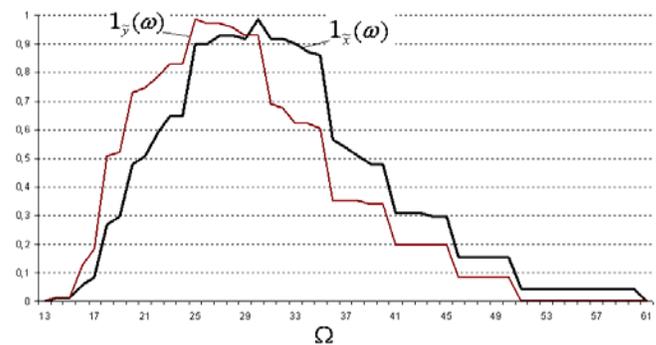

Fig.1. Eventological membership functions of vague events $\widetilde{x}$ - "*the young man*" and $\widetilde{y}$ - "*the young woman*", constructed as a result of vague experiment in which participated $|M| = 71$ reasonable subjects. It is necessary to notice, that membership functions of vague sets, constructed by a method of expert estimations, under certain conditions coincide with the indicator of corresponding vague event.

Thus, we formalized atomic syntagmas "*the young man*" and "*the young woman*" by

E-linguistic variable $L: =$ "*Age*" by vague events $\widetilde{x}$ and $\widetilde{y}$.

Links, uncertainty and the negation defined by the syntactic procedure $G_E$, are formalized by means of set-operations above set of *M*-vague events $\widetilde{X} = \{\widetilde{x}, x \in X\}$, generated by the set $X$ [4] according to Minkovsky. The basic eventological theorem about vague events [4] offers the formula for calculation of the indicator (eventological membership functions) of any set-operation above set of vague. Let's consider a syntagma "*the young man and the young woman*", constructed according to the syntactic rule $G_E$. We formalize a link "*and*" as intersection according to Minkovsky of vague events $\widetilde{x}$ and $\widetilde{y}$: with eventological membership function $\mathbf{1}_{\widetilde{x}(\cap)\widetilde{y}}(\omega)$, the satisfying formula for the indicator of any set-operation above the set of $M$-vague events $\widetilde{X}$ (fig. 2).

In Gau and Buehrer's theory language links cannot be mathematically interpreted only by one type of conjunction in all situations. For example, the link "*and*" is presented by the t-norms, being special binary operations on an interval [0, 1]. The choice of a concrete kind of the formula for a link depends on the mutual attitude between vague sets that leads, thus, to use of various t-norms [3]: a minimum $T_M$, probabilistic product $T_P$ and Lukasiewicz's t-norm (fig. 2), etc. The basic eventological theorem of vague events [4] offers the one and only general the eventologically correct formula for the indicator of any set-operation above any set of vague events. Thus, the formula for eventological membership functions of E-event to result of the given set-operation is unique eventologically correct generalization of "usual" membership functions, various empirical variants used in set in the Gau and Buehrer's theory of vague sets. The basic theorem has the general character as any set-operation habitual union, intersection and a symmetric difference can act, and also any other possible set-operation above set of vague events. Because of absence of the similar theorem in the Gau and Buehrer's theory of vague sets the set of variants of membership function to the same set-operations till now is used: actually, to each specific task the variant of membership function to this or that set-operation is searched. This essential lack managed to be avoided in the offered eventological theory of vague events. Besides this the eventology specifies, that the structures of dependences of "usual" events of which vague events consist, serve as the reason of plurality of variants of membership function in the Gau and Buehrer's theory of vague sets. Only structures of dependences of events define a kind of eventological membership function. And a plurality of variants in the classical theory speaks that it is necessary to lean not only on membership functions, which do not bear the information on vague events, and on all eventological distribution of set of events which make the given vague event, and on eventological distribution of set of vague events as it is suggested in new the eventological theory of vague events.

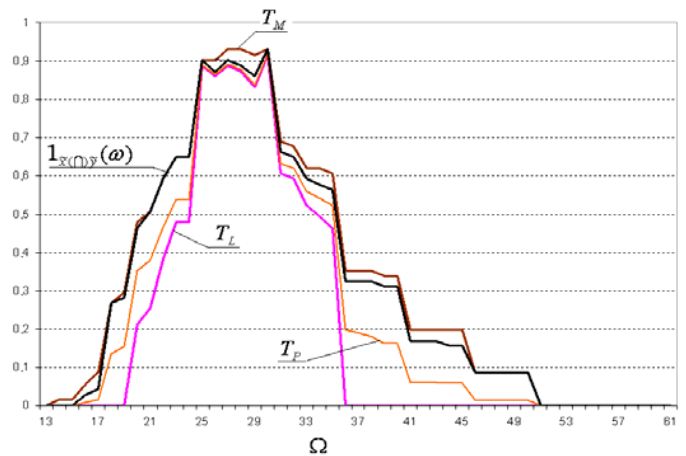

Fig. 2. Various interpretations of a syntagma "the *youngman and the young woman*".

The link "*and*" is presented by the indicator of intersection of two vague events o according to Minkovsky $\mathbf{1}_{\widetilde{x}(\cap)\widetilde{y}}(\omega)$ and the most popular t-norms [3]: a minimum $T_M$, probabilistic product and Lukasiewicz's tnorm. However this example allows illustrating evidently the basic concepts and principles of the theory of vague events.

## 3. CONCLUSION

The theory of vague events is one of directions of the eventology [4], convincingly showing efficiency of eventological theory in knowledge of the phenomena and processes where the leaging role is played by the reasonable subject. The eventological theory of vague events naturally mathematically proves and expands the Gau and Buehrer's theory of vague sets as the approach in the mathematical description of uncertainty. In this work very simple example of eventological formalizations of a linguistic variable is considered.

**Supriya Raheja ,Assistant Professor in ITM University**.She had done her engineering from Hindu college of Engineering,Sonepat and masters from Guru Jambeshwar University of Science and Technology,Hisar.

**Smita Rajpal,**ITM University .She is a researcher in the field of Soft Computing.Published various papers in International Journals and Conferences.Also serving many International journals and conferences as a reviewer/ committee member /Editrorial Board member.